\documentclass{article}
\usepackage{ijcai26}

\usepackage{times}
\usepackage{soul}
\usepackage{url}
\usepackage[hidelinks]{hyperref}
\usepackage[utf8]{inputenc}
\usepackage[small]{caption}
\usepackage{graphicx}
\usepackage{amsmath}
\usepackage{amsthm}
\usepackage{booktabs}
\usepackage{algorithm}
\usepackage{algorithmic}
\usepackage[switch]{lineno}

\title{MixINN: Accelerating Plant Breeding by Combining Mixed Models and Deep Learning for Interaction Prediction}

\iftrue
\author{
Aike Potze$^1$
\and
Fred van Eeuwijk$^2$\and
Ioannis N. Athanasiadis$^{1}$\\
\affiliations
$^1$Artificial Intelligence Group, Wageningen University \& Research\\
$^2$Mathematical \& Statistical Methods Group (Biometris), Wageningen University \& Research\\
\emails
\{aike.potze, fred.vaneeuwijk, ioannis.athanasiadis\}@wur.nl
}
\fi

\usepackage{amsmath}
\usepackage{amsfonts}
\usepackage{booktabs}
\usepackage{multirow}
\usepackage{array}
\usepackage{calc}
\usepackage{tabularx}
\usepackage{xcolor}
\usepackage{enumitem}

\newcolumntype{C}[1]{>{\centering\arraybackslash}m{#1}}

\newif\ifrevision
\revisionfalse 

\begin{document}

\maketitle

\begin{abstract}
    Plant breeding underpins global food security through incremental, accumulating improvements in crop yield, quality and sustainability, achieved via repeated cycles of crop ranking, selection and crossing. Climate change disrupts this process by altering local growing conditions, thereby shifting the relative performance of crop genotypes. Predicting these relative changes in yield is critical for food security. Yet, this problem remains an open challenge in plant breeding, and relatively unexplored within the AI community. We propose MixINN, an approach that first isolates high-quality genotype-environment interaction labels using mixed models, and then predicts these interactions for new crop varieties in future environmental conditions with a deep neural network. We evaluate our method on a corn multi-environment trial across the continental United States and show improved prediction of genotype ranking over current plant breeding methods. MixINN demonstrated superior performance in identifying the 20\% most productive corn genotypes, leading to a 5.8\% higher average yield, which further improved to 7.2\% when targeting specific growing environments. These are competitive results for real-world breeding programs, demonstrating the potential of AI research in accelerating the development of climate-adapted crops, and improving future food security under climate change.
\end{abstract}

\section{Introduction}
Plant breeding forms the foundation of global food security through the continuous improvement of crop genetic potential. Over the past decades, breeding is responsible for an estimated yearly increase in potential yield of 0.6\%-2\% across staple crops~\cite{fischer2014crop}. Through repeated selection of genotypes for desirable traits, breeders improve the genetic architecture underlying these traits, which enhances genotype performance across breeding cycles. This genetic improvement, referred to as \textit{genetic gain}, accumulates over successive cycles, leading to substantial long-term advances in crop productivity~\cite{ramakers2025evaluation}.
Crucially, genetic gain per selection cycle is proportional to the prediction accuracy in breeding trials~\cite{xu2017enhancing}. Subsequently, improving prediction accuracy directly raises future crop yields, and is critical for future food security.

Plant breeders can increase prediction accuracy by using a combination of designed experiments and modelling strategies that explicitly express the dependence of phenotypic traits on both genetic and environmental inputs. The experimental design of individual trials and large multi-environment trial (MET) networks induce structured replication of genotypes across environments. This structured replication is then modelled with mixed models~\cite{malosetti2013statistical}. Mixed models, also known as hierarchical or multilevel models, capture structured replication through explicit assumptions on variance-covariance structures~\cite{elias2016half}. Jointly, advanced experimental designs and mixed models enable breeders to disentangle genetic signal from environmental noise and natural variability, and consequently improve genetic gain per breeding cycle.

Crucially, the relative performance of genotypes is conditional on specific environments, a phenomenon named genotype-by-environment interaction (G×E). This affects prediction of genotype performance when growing environments are different from selection environments~\cite{romagosa2013genotype}. Climate change is increasing this misalignment, potentially reducing genetic gain \cite{cooper2023extending}, and threatening future food security. Selection can be adjusted to specific environments through explicit prediction of G×E from genetic markers and environmental variables. As such, predictive models for G×E improve genetic gain under changing environments. To spur the development of such models, plant breeding researchers across the United States have recently collected and shared the first large-scale open MET dataset~\cite{lima2023genomes}, with 4,683 of corn genotypes grown across 243 unique environments in the US corn belt, during 2014-2022. The researchers subsequently organized a competition with these data, which enabled the first large-scale intercomparison of predictive models for the prediction of G×E~\cite{washburn2025global}.

Current statistical models assume linear relationships of yield with environmental variables, despite strong evidence that the physiological mechanisms underlying G×E are inherently nonlinear \cite{napier2023gene}. This has motivated the development of nonlinear approaches, including the application of deep learning-based methods. However, deep learning has been largely unsuccessful thus far in large-scale intercomparison studies: the recent intercomparison results~\cite{washburn2025global} showed no advantage of deep learning-based methods over alternatives, and revealed that best-performing models are based on ensembles of linear mixed models~\cite{meuwissen2001prediction,jarquin2014reaction} and classical machine learning algorithms such as Random Forests~\cite{breiman2001random}.

Recently, \citeauthor{potze2025}~(\citeyear{potze2025}) suggested that neural networks under-utilize genetic signals due to \textit{unimodal bias} \cite{cadene2019}, where one dominant modality of data disrupts the learning of another modality. The authors proposed to isolate genetic and environmental effects from G×E with a fixed effects model, and predict them separately with deep learning models. Their proposed method, SINN, enhanced utilization of genetic information and outperformed previous models in ~\cite{washburn2025global}. However, SINN used interaction residuals from a two-way fixed main effects Analysis of Variance (ANOVA) model as labels for G×E, leading to a poor signal-to-noise ratio in training samples and limited success in the prediction of G×E~\cite{potze2025}.

We argue that this limitation is directly addressed by the mixed models used by breeders, as they are designed to separate genetic signals from structured noise. We therefore propose MixINN, the Mixed-model Interaction Neural Network, as an approach to integrate mixed models with deep learning models to improve the prediction of GxE. First, a factor-analytic mixed model is fitted to decompose training samples into environment corrected genetic effects of low dimension, environmental effects, interaction effects and noise. Next, individual deep learning models were trained to predict genetic, environmental and interaction effects for new genotypes and environments. MixINN leverages explicit assumptions about correlation structure induced by the experimental design, aiming to isolate high-quality genotype and GxE effects, which consequently improve downstream ranking prediction by the neural network.

This work combines the contribution of stakeholders with a diverse set of backgrounds, including plant breeders and quantitative geneticists. The statistical assumptions in MixINN were based on the experience of quantitative geneticists directly involved with the work. The choice of representative selection strategies and scenarios to assess real-world impact was informed by plant breeders. Furthermore, this work is indirectly enabled by the growers, breeders and researchers of the Genomes To Fields Initiative, through the creation of the open MET dataset used in this study~\cite{lima2023genomes}.

We demonstrate that MixINN surpasses current state-of-the-art plant breeding models in identifying high-yielding crop genotypes in future environments. We then show that this directly translates to the selection of higher-yielding corn varieties in real-world data, and conditions mirroring operational breeding. Our results indicate that MixINN can accelerate crop yield improvements and the selection of crops adapted to future environmental conditions, and contribute towards future food security under climate change.

To summarize, in this work we:
\begin{itemize}[topsep=0pt,leftmargin=9pt,itemsep=0pt
 ,partopsep=0pt,parsep=0pt]
\item Introduce the prediction of environment-specific genotype performance as an open problem in plant breeding
\item Propose MixINN, a method that combines mixed models from plant breeding experts with neural networks
\item Improve prediction accuracy over current state-of-the-art methods with MixINN
\item Demonstrate gains in crop yield estimation performance and improved selection for environmental adaptation in real-world breeding data
\end{itemize}

\section{Related Work}
\subsubsection{Prediction of G×E}
In the domain of plant breeding, classical statistical methods for G×E include linear-bilinear methods~\cite{finlay1963analysis,gauch1992statistical,yan2002gge} and linear mixed models~\cite{piepho1997analyzing,smith2001analyzing}. Seminal work by~\citeauthor{bernardo1994prediction} (\citeyear{bernardo1994prediction}) formulated GBLUP, which integrated genetic markers into linear mixed models and enabled prediction for new genotypes. This work was extended with environment and interaction kernels by ~\citeauthor{jarquin2014reaction} (\citeyear{jarquin2014reaction}), supporting G×E prediction. Current approaches are centered around extensions of this work in linear mixed models~\cite{hu2025megalmm}, process-based models~\cite{cooper2016use} and machine learning methods~\cite{crossa2025machine}. Modern deep learning-based methods have shown mixed results across case studies~\cite{khaki2019crop,washburn2021predicting,kick2023yield}. Recently, SINN~\cite{potze2025}, attributed poor performance to \textit{unimodal bias}~\cite{cadene2019} and addressed it by predicting the marginal effects of genotype, environment and G×E separately. However, the naive decomposition biased estimates of effects, and did not separate G×E from noise, limiting prediction accuracy for G×E.

\section{Methodology}
\subsection{Problem formulation}
Let \mbox{$I = \{1, \dots, n_g\}$} denote the set of crop genotypes and \mbox{$J = \{1, \dots, n_e\}$} denote the set of environments. We aim to predict crop yield, which is formalized as a function 
\begin{equation}
    f: I \times J \to \mathbb{R},
\end{equation}
where continuous target values $y_{ij}\in \mathbb{R}_{\geq 0}$, representing crop yield, are estimated for any genotype-environment combination $(i,j)$. To enable the estimation of $y_{ij}$ for new genotypes and new environments, we define two sets of side features. For each genotype $i$, we define a genetic feature vector $x^g_i \in \mathbb{R}^{d_g}$ consisting of $d_g$ genetic markers or features. For each environment $j$, we define an environmental feature vector $x^e_j \in \mathbb{R}^{d_e}$, consisting of $d_e$ weather, soil and farm management features. Given a set of $n_s$ training samples in $\mathcal{S} = \{(x^g_i, x^e_j, y_{ij})\}$, the estimation of crop yield $y_{ij}\in \mathbb{R}_{\geq 0}$ can then be formulated as the mapping $f$ of combinations of feature vectors $(x^g_i, x^e_j)$ to yield $y_{ij} = f(x^g_i, x^e_j)$.

\subsection{Dataset}
We implement our methods using the Genomes to Fields 2022 Maize G×E Prediction Challenge \cite{lima2023genomes} dataset. This dataset is the result of a long-running collaboration between plant breeding researchers across the United States, and consists of over 140,000 records of corn yield from 2014-2022. After filtering and imputation (Appendix A) we retain 123,517 training samples, representing 4,417 genotypes grown across 212 environments. With each record, we include: (i) 20,000 genetic markers per genotype, encoded as $\{-1, 0, 1\}$ representing a homozygous reference marker genotype, heterozygous marker genotype, or homozygous alternate marker genotype at each position. (ii) 11 daily weather features, averaged over the growing season, (iii) 20 soil features, and (iv) 2 management features. We concatenate weather, soil and management features into an environmental feature vector of length 33.
We follow the experimental setup of \cite{washburn2025global}, which excludes all yields of the year 2022 as hold out test samples. The test set covers two scenarios: new environments (nE), and new genotypes and new environments (nGE). In total, the test set consists of over 11,556 samples, corresponding to 6\% nE and 94\% nGE. The test set environments are repeated locations from the training set, with new weather conditions.

\subsection{Proposed method} 
In this section we introduce MixINN: a two-step approach integrating mixed models with deep neural networks. First, we propose to correct and decompose observed yields with linear mixed models: a class of models widely utilized in plant breeding due to their superior ability to correct performance of genotypes within trials through modelling of variance-covariance (VCOV) structures~\cite{robinson1991blup,piepho2008blup}. Next, using a factor-analytic mixed model, we correct observed yield values and estimate genotype effects, environment effects and G×E effects. These effects are then used as label sets for for the structured optimization of a two-tower neural network architecture~\cite{covington2016deep} to predict GxE.

\subsubsection{Factor-analytic modelling of covariance}
The MixINN approach starts with fitting the following linear mixed model to $\mathcal{S}$ using restricted maximum likelihood \cite{patterson1971recovery}:
\begin{equation}\label{eq:decomp}
    y_{ijk} = \mu + G_i + E_j + GE_{ij} + \epsilon_{ijk},
\end{equation}
\noindent
where $y_{ijk}$ is the observed yield for replicate $k$ of genotype $i$ in environment $j$, $\mu$ is the overall mean, $G_i$ is the random effect of genotype $i$, $E_j$ is the fixed effect of environment $j$,  $GE_{ij}$ the random effect of genotype $i$ in environment $j$, and $\epsilon_{ijk}$ the residual of replicate $k$. 

Following established best practices on variance-covariance (VCOV) modelling in plant breeding trials~\cite{boer2007mixed,piepho2024factor} designed to isolate genetic signals from environmental signals and structured noise, we introduce three assumptions: (i) genotype effects are independently sampled from a common population of genotypes, with VCOV structure for the genotypes \mbox{$\boldsymbol{\Sigma}_g \in \mathbb{R}^{n_g \times n_g}$} defined as \mbox{$\boldsymbol{\Sigma}_g = \sigma_g^2 \mathbf{I}_g$}. (ii) Genotype-environment interactions $GE_{ij}$ are sampled across a population of genotypes, and a population of environments. As such, we model interactions as random interactions of genotypes and environments, with joint VCOV structure expressed as the Kronecker product between $I_g$ and VCOV structure \mbox{$\boldsymbol{\Sigma}_e \in \mathbb{R}^{n_e \times n_e}$} for the environments.
(iii) Genetic features map directly to genotype, while environmental features have variation within environments that is not expressed in $x^e_j$. As such, we assume each environment has a unique residual variance $\sigma_{\epsilon,j}^2$, expressed as block diagonal VCOV matrix $\boldsymbol{\Sigma}_\epsilon$. We specify these assumptions as:
\begin{subequations}
\begin{align}
\mathrm{G}_{i} \sim \mathcal{N}&\left(0,\boldsymbol{\Sigma}_g\right),\;\;\;\;\;\;\;\;\;\;\;\;\,\,  \boldsymbol{\Sigma}_g =\sigma_g^2\mathbf{I}_g,\\
\mathrm{GE}_{ij} \sim \mathcal{N}&\left(0,\mathbf{I}_g\otimes\boldsymbol{\Sigma}_e \right),\quad\:\:\: \boldsymbol{\Sigma}_e = \boldsymbol{\Lambda} \boldsymbol{\Lambda}^\top + \boldsymbol{\Psi},\\
\epsilon_{ijk} \sim \mathcal{N}&\left(0,\boldsymbol{\Sigma}_\epsilon\right), \quad\quad\quad\;\;\; \boldsymbol{\Sigma}_\epsilon = \mathrm{diag}(\sigma_{\epsilon,j}^2\mathbf{I}_j),
\end{align}
\end{subequations}
\noindent directly estimating $\boldsymbol{\Sigma}_e$ would require the estimation of \mbox{$0.5n_e(n_e+1)$} variance and covariance parameters, rendering the approach computationally infeasible. Instead, we approximate $\boldsymbol{\Sigma}_e$ with $\boldsymbol{\Lambda}$ and $\boldsymbol{\Psi}$, an approach named factor-analytic (FA) VCOV estimation~\cite{piepho1998empirical}. In this formulation, \mbox{$\boldsymbol{\Lambda} \in \mathbb{R}^{n_e\times r}$} is matrix of $r$ factor loadings per environment, and \mbox{$\boldsymbol{\Psi}\in\mathbb{R}^{n_e\times n_e}$} is a diagonal matrix capturing environment-specific variances. We fit two latent factors ($r=2$) and one variance parameter per environment, resulting in $3n_e-1$ parameters used to estimate $\boldsymbol{\Sigma}_e$.

Next, from the fitted model \eqref{eq:decomp} we obtain corrected yield $\hat{y}_{ij}$ and marginal expectations over $I$ and $J$, approximate $\hat{\mu}$ and generate label sets for genetic effects ($y^g_i$), environment effects ($y^e_j$), and G×E ($y^{ge}_{ij}$):
\begin{subequations} \label{eq:est}
    \begin{align}
        \hat{\mu} &= \frac{1}{n_s} \sum_{s=1}^{n_s} y_{s},\quad\quad\quad\quad\; y^g_i= \frac{1}{n_e} \sum_{j=1}^{n_e} \hat{y}_{ij} - \hat{\mu},\label{eq:fdec:three} \\
        y^e_j&= \frac{1}{n_g} \sum_{i=1}^{n_g} \hat{y}_{ij} - \hat{\mu},\label{eq:fdec:four} \quad\quad y^{ge}_{ij} = \hat{y}_{ij} - \hat{\mu} - y^g_i - y^e_j,
    \end{align}
\end{subequations}

\subsubsection{Structured optimization}
We construct a two-tower neural network, which is a commonly utilized architecture for predicting interactions in content-based recommendation~\cite{covington2016deep}. It contains a Multilayer Perceptron (MLP~\cite{rumelhart1986learning}) based genotype encoder, an MLP-based environment encoder, and a dot-product fusion module. To optimize the network for generalization to unobserved genotypes and environments, we follow a stage-wise, structured optimization scheme~\cite{potze2025}. We start by defining three component tasks and functions
\begin{subequations} \label{eq:modules:all}
    \begin{align}
        y^g_i &= f_g(x^{g}_i), \label{eq:modules:one}  \\
        y^e_j &= f_e(x^{e}_j), \label{eq:modules:two} \\
        y^{ge}_{ij} &= f_{ge}(x^{g}_i, x^{e}_j), \label{eq:modules:three}
    \end{align}
\end{subequations}
\noindent
and take the genetic encoder as $f_g$, environment encoder as $f_e$, and the full two-tower network as $f_{ge}$. We use the same loss function across component tasks and models:
\begin{equation} 
    \min_{f_\ast} \; \mathcal{L} = \ell_2(f_\ast(x^\ast), y^\ast), \quad \text{for } \ast = g, e, ge.
\end{equation}

After training $f_g$ and $f_e$ on respectively $y^g_i$ and $y^e_j$, we use them to initialize the encoders of $f_{ge}$. As final step, we combine $\hat{\mu}$, $\hat{f}_g$, $\hat{f}_e$ and $\hat{f}_{ge}$:

\begin{equation}
    \hat{y}_{i_mj_m} = \hat{\mu} + \hat{f}_g(x^g_{i_m}) + \hat{f}_e(x^e_{j_m}) + \hat{f}_{ge}(x^g_{i_m}, x^e_{j_m}),\label{eq:recomposition}
\end{equation}
to approximate $y$ for sample $m$ with unobserved genotypes and environments.
\subsubsection{Implementation details}
As genetic encoder we use an MLP that takes a feature vector of 20,000 genetic markers as input, with two hidden layers and one linear output layer, trained on $y^g$. We use Layer Normalization~\cite{ba2016layer} and Dropout~\cite{srivastava2014dropout} after each hidden layer. As activation function we use ReLu~\cite{nair2010rectified}. We halve the number of nodes in the final hidden layer and use sigmoid as activation function, to encourage informative embeddings. We set the dropout ratio to 0.5 across all layers, set batch size to 256 and train for 250 epochs with an AdamW optimizer~\cite{loshchilov2019decoupled}, and a Mean Squared Error (MSE) loss. After hyperparameter tuning (Appendix B), we train an MLP with 256 nodes per layer, at a learning rate of $1\times10^{-3}$ and a weight decay of $3\times10^{-4}$.

We use the same MLP structure as an environment encoder, which maps the environment feature vector to $y^e$. After hyperparameter tuning, we obtain an MLP with three hidden layers and 48 nodes per layer, which we train for 500 epochs, at a batch size of 32, a learning rate of $1\times10^{-3}$ and a weight decay of $1\times10^{-5}$. 

When training the full two-tower architecture on $y^{ge}$, we discard the output layer of the genetic and environment encoders and use the final feature vectors as genotype and environment embeddings. Our interaction module consists of one linear layer per encoder, which projects the embeddings to a common embedding length. Finally, both projected embeddings are fused with a dot product into a single output value. The full two-tower architecture is initialized with $\hat{f}_g$ and $\hat{f}_e$, and trained for 250 epochs, at a batch size of 256, a learning rate of $1\times10^{-2}$ and a weight decay of $3\times10^{-4}$. We use an embedding length of 8.

\subsection{Baseline models}
We consider two groups of baselines to compare with. As first group we include the top ten best-performing models in terms of RMSE from the original intercomparison study~\cite{washburn2025global}. As the second group, we manually implement a range of domain-specific models, including one kernel-based method using only genetic information (GBLUP~\cite{bernardo1994prediction,meuwissen2001prediction}) and one kernel-based method that includes environmental information and interactions (G×EBLUP~\cite{jarquin2014reaction}, following implementation in~\cite{potze2025}. The memory requirements of G×EBLUP scale quadratically with dataset size. As we were constrained to 500GB of RAM, we were unable to fit G×EBLUP on the full dataset. We removed replicates and random samples from the most common genotypes until it was possible to fit the model. We were able to fit G×EBLUP with 40\% of samples. For completeness, we include the exact implementation of the models in Appendix C.

As additional baselines in the second group, we include one deep learning-based method developed on the Genomes to Fields dataset~\cite{kick2023yield}, denoted as G2F-DNN in this work. G2F-DNN follows an intermediate fusion approach, with a Convolutional Neural Network~\cite{lecun1989backpropagation} as encoder for daily weather features, an MLP as encoder for soil and management data, an MLP operating on principal components as genetic encoder, and an MLP as interaction network. We tune learning rate and weight decay \mbox{(Appendix B)}. We also include the original SINN model as additional baseline. As it was developed using the exact same dataset, we retain the original hyperparameters.

Due to year-by-year variation, optimization of each model was highly sensitive to the specific choice of validation set. As such, all models are fitted on eight cross-validation folds, each consisting of one holdout year and 12.5\% of total training genotypes. Models predicting $y^g$ were evaluated on holdout genotypes, models predicting $y^e$ on holdout environments and models predicting $y^{ge}$ or $y$ on their intersection. The eighth cross-validation fold (corresponding to the year 2021) was used for hyper-parameter tuning. All reported results for all models are from predictions averaged across eight training folds. This process is replicated ten times to calculate standard deviations and significance, resulting in total of 80 models evaluated per model type.

\begin{table*}[t!]
\centering
\begin{tabular}{l|p{0.4cm}p{0.4cm}p{0.4cm}|p{1.9cm}p{1.9cm}|p{1.9cm}p{1.9cm}p{1.9cm}}
\toprule
\textbf{Name / Model} &  
\multicolumn{3}{c|}{\textbf{Type}} &
\multicolumn{2}{c|}{\textbf{Ranking}} & 
\multicolumn{3}{c}{\textbf{Regression}}\\
&ST&ML&DL&
\multicolumn{1}{c}{ $r_j \uparrow $} & \multicolumn{1}{c|}{ $\rho_j \uparrow $ } & \multicolumn{1}{c}{ RMSE    $\downarrow  $} & \multicolumn{1}{c}{ MAE  $\downarrow $ } & \multicolumn{1}{c}{ $r \uparrow  $ } \\
\midrule
\midrule
AIBreeding & \checkmark & \checkmark & \checkmark & 0.22 & 0.14 & 2.76 & 2.15 & 0.44 \\
MPB\_Group & \checkmark &  & \checkmark & 0.26 & 0.18 & 2.74 & 2.12 & 0.49 \\
ML\_APT & \checkmark & \checkmark &  & 0.19 & 0.10 & 2.60 & 2.06 & 0.56 \\
SmAL &  &  & \checkmark & 0.15 & 0.12 & 2.52 & 2.00 & 0.59 \\
Purdue & \checkmark & \checkmark &  & 0.16 & 0.12 & 2.49 & 2.00 & \textbf{0.63} \\
CGM & \checkmark &  &  & 0.35 & 0.28 & 2.49 & 1.98 & 0.59 \\
UCD\_MegaLMM & \checkmark &  &  & 0.34 & 0.23 & 2.50 & 1.96 & 0.62 \\
phenomaize & \checkmark &  &  & 0.24 & 0.19 & 2.47 & 1.97 & 0.62 \\
igorkf &  & \checkmark &  & -- & -- & 2.46 & 1.94 & 0.60 \\
CLAC & \checkmark & \checkmark &  & 0.36 & 0.31 & 2.46 & 1.93 & \textbf{0.63} \\
\midrule
GBLUP & \checkmark &  &  & 0.26 $\pm$ 0.00 & 0.16 $\pm$ 0.00 & 3.05 $\pm$ 0.01 & 2.49 $\pm$ 0.01 & 0.16 $\pm$ 0.00 \\
G×EBLUP & \checkmark &  &  & 0.38 $\pm$ 0.00 & 0.31 $\pm$ 0.01 & 2.90 $\pm$ 0.01 & 2.35 $\pm$ 0.01 & 0.43 $\pm$ 0.00 \\
G2F-DNN &  &  & \checkmark & 0.25 $\pm$ 0.02 & 0.17 $\pm$ 0.02 & 2.85 $\pm$ 0.08 & 2.29 $\pm$ 0.09 & 0.47 $\pm$ 0.03 \\
SINN & \checkmark &  & \checkmark & 0.38 $\pm$ 0.00 & 0.30 $\pm$ 0.01 & \textbf{2.40} $\mathbf{\pm}$ \textbf{0.04*} & \textbf{1.85} $\mathbf{\pm}$ \textbf{0.03} & \textbf{0.63} $\mathbf{\pm}$ \textbf{0.02} \\
\midrule
MixINN (ours) & \checkmark &  & \checkmark & \textbf{0.41} $\mathbf{\pm}$ \textbf{0.00*} & \textbf{0.36} $\mathbf{\pm}$ \textbf{0.00*} & 2.46 $\pm$ 0.03 & 1.87 $\pm$ 0.02 & 0.61 $\pm$ 0.01 \\
\bottomrule
\end{tabular}
\caption{Prediction performance of (top) models from literature, (middle) implemented baselines and (bottom) MixINN (this work). Model types include statistical models (ST), machine learning (ML) and deep learning (DL). Ranking metrics are measured as average correlation within each environment. RMSE and MAE are in Mg/ha. Best model per column is shown in \textbf{bold}, with * denoting $p < 0.05$ (independent t-test of best model with implemented models, and single-sample t-test against results from literature).}
\label{tab:g2f}
\end{table*}

\subsection{Evaluation}
We evaluate MixINN on two aspects: (1) predictive performance compared to current state-of-the-art in plant breeding, and (2) actual breeding outcomes. 

For evaluating predictive performance we follow the exact test set of~\cite{washburn2025global}, and include the same regression metrics including Root Mean Squared Error (RMSE), Mean Absolute Error (MAE) and Pearson correlation coefficient $r$. As ranking metrics, We include within-environment Pearson correlation $r_j$ and Spearman's rank correlation coefficient $\rho_j$, calculated within each environment and averaged across environments.

We quantify breeding outcomes by emulating selection based on predicted yields. In a breeding program, only a fixed top-performing percentage of genotypes is selected in each cycle. To emulate this, we subset the test data based on predicted performance of individual genotypes. Then, we evaluate the average yield of the selected genotypes. The more accurate the predicted rankings, the higher the average yield of selected genotypes and the greater the expected genetic gain. Importantly, this procedure emulates selection by applying model-based rankings to an existing test set, without evaluating subsequent generations. Accordingly, reported gains reflect selection response within the test set rather than cumulative gain across breeding cycles.

We select based on global ranking of genotypes. To avoid distortion of rankings due confounding between genotypes and specific environments, we exclude genotypes from global selection which are missing in more than 10\% of environments of the test set. This leaves us with 65\% of genotypes to select from, with an average yield lowered by 0.1 Mg/ha. Finally, we fit the following linear mixed model to the predicted yields of each model:
\begin{equation}
    y_{ij} = \mu_j + G_i + \epsilon_{ijk} \quad\quad G_i \sim \mathcal{N}\left(0,\sigma^2_g\right),
\end{equation}
\noindent
and rank genotypes according to the extracted random genetic effect $G_i$ across predictions.

We also evaluate a second selection strategy: selection based on the specific environmental conditions in each test environment. This measures capacity of the models to adapt selection decision to specific environmental conditions. For this, we again predict the yield for each genotype-environment combination, and retain the top genotypes per environment based on the proportion of genotypes selected.

\section{Experiments}
In this section, we answer two key questions, in order of increasing practical relevance to breeding:
\begin{enumerate}[topsep=-1pt,itemsep=-1ex,partopsep=1ex,parsep=1ex]
  \item How do models perform compared to the current state-of-the-art in plant breeding?
  \item How do better ranking predictions increase crop yields across selection strategies?
\end{enumerate}
\begin{figure*}[tb!]
    \centering
    \includegraphics[width=1.0\textwidth]{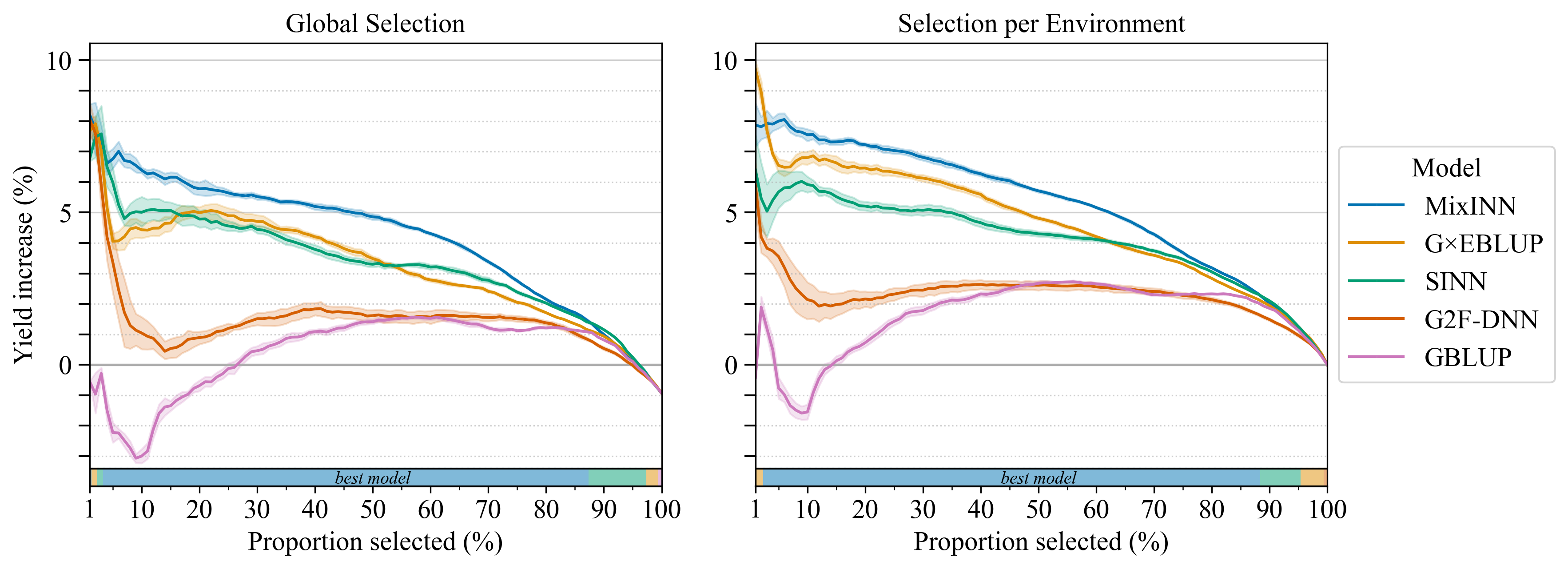}
    \caption{Increase in average plot yields when only corn genotypes with highest predicted yields are grown, plotted versus percentage selected across test set, with selection based on global prediction of genotype performance (left) and environment-specific prediction of genotype performance (right). Shaded areas denote 95\% confidence interval, and best model per selection percentage is shown below. Average yield of entire test set (10.1 Mg/ha) was used as reference.}
    \label{fig:results:yield}
\end{figure*}
\subsubsection{1. Comparison with plant breeding models} The first experiment exactly reproduces the setup of \citeauthor{washburn2025global}~(\citeyear{washburn2025global}), which measures the ability of models to predict yield across the next generation of a multi-environment trial. The test set is all new environments, and mixes known and new genotypes.
Table \ref{tab:g2f} shows the performance of MixINN against both sets of baseline models. 

We find that MixINN significantly improves over current plant breeding models across ranking metrics. Previously, the state-of-the-art performance in ranking prediction was achieved by SINN and G×EBLUP. MixINN surpasses them with a relative improvement of about 10\% in $r_j$ and 20\% in $\rho_j$. The top five models in terms of ranking performance (MixINN, SINN, G×EBLUP, CLAC, CGM) all utilized linear or mixed models (ST). Purely deep learning-based (G2F-DNN, SmAL) and machine-learning based models (igorkf) demonstrated poor performance in ranking prediction.

For the regression metrics, only SINN demonstrates a significant advantage over MixINN in terms of RMSE. All other models achieved comparable or lower performance. We note that deep learning models (AIBreeding, MPB\_Group, SmAL, G2F-DNN) perform poorly on regression metrics, with the exception of those utilizing structured training (MixINN, SINN). Both statistical models and machine learning models perform well on regression metrics.

We find that ranking and regression metrics do not directly align. Ranking metrics measure correlation within environments, and are invariant to prediction of environmental means. As such, we infer that models with good regression and poor ranking performance poorly predict genetic variation, emphasizing the prediction of environmental means instead. Subsequently, ranking performance is more relevant to breeding than regression performance, and one can expect higher gains in yield when selecting genotypes based on the best-performing model in ranking prediction.

\subsubsection{2. Impact on yield of selected genotypes}
To fully understand how each model impacts breeding outcomes, we consider the outcomes of selecting genotypes in the test set according to the predictions of each model. We consider two selection strategies: selection based on predicted mean yield of genotypes across all test environments (global selection), and selection based on predicted yield of genotypes within each specific test environment (selection per environment). Increase in average yields per selection strategy and fraction are shown in Figure \ref{fig:results:yield}.

MixINN achieves a significant improvement in yield increase over the other models, across small to large selection proportions of 5\% - 50\% of genotypes, and across both global and environment-adapted selection strategies. G×EBLUP performs second on selection per environment, and both SINN and G×EBLUP have comparable performance on global selection. Neither G2F-DNN or GBLUP increase yield consistently, with GBLUP even decreasing yield with global selection. We attribute this to confounding of genetic and environmental signals in the training set, which cannot be resolved without including environmental features or statistical factors absorbing environmental variation. When these are added to GBLUP, this issue vanishes (G×EBLUP).

At very low proportions of selection the variance between replicates increases and  relative performance of each model type shifts. For the global selection, all models except GBLUP perform equally well at selection proportions below 5\%. For local adaptation, G×EBLUP has the highest average yield at proportions below 5\%. Selection for local adaptation achieves a higher average yield across selection intensities, and the relative performance of models is consistent across selection intensities and selection strategies. 

We quantify impact of MixINN on future food security based on the increase in yield achieved by selection under conditions representing operational breeding. As representative conditions, we apply global selection of 20\% of genotypes. MixINN improved yield by 5.8\%, corresponding to a relative increase over the second-best model (G×EBLUP) of 15.8\%. This suggests that the yearly gains in crop yield achieved by breeding programs implementing MixINN could accelerate by up to 15.8\% compared to current yearly gains.

Similarly, we quantify the impact of MixINN on environmental adaptation of breeding programs based on selection per environment under the same selection proportion. MixINN increased yield by 7.2\%, improving over the second-best model (G×EBLUP) by 11.9\%. This demonstrates that yield gains can be further accelerated by targeting specific environmental conditions, and that MixINN is the best suited for this. However, we note that costs rise with the number of spatial units selected for. Breeders therefore generally target groups of similar environments, with granularity depending on exact breeding context. However, our results show that MixINN significantly improves yield gains over other models across both ends of this spectrum of environmental adaptation.

\section{Discussion}
\subsection{Accurate ranking raises crop yields}
In this work, we introduced the prediction of environment-specific genotype performance as a critical open problem in plant breeding. We evaluated a selection of domain-specific models on a comprehensive, real-world dataset, with field trials spanning the corn-growing regions of the United States, across 9 years, and 4,683 unique corn varieties. This dataset enabled us to directly measure the impact of each model in real-world plant breeding contexts, spanning both global and environment-specific selection, and a wide range of selection intensities.

The proposed method, MixINN, outperformed previous domain-specific statistical models and machine learning models in terms of correct rankings of genotypes and increased gain in crop yields. We observe that this comes at a cost of regression-based metrics. However, we find that regression-based metrics do not translate to gains in real-world yield of selected genotypes, underscoring the importance of aligning evaluation with breeding impact, rather than point-based predictive accuracy.

The results highlight that the social impact of new predictive models in plant breeding hinges on the active participation of the breeders. The choice of target metrics should be informed by breeding outcomes: improving regression metrics does not necessarily improve yield of selected genotypes, while ranking metrics are more closely aligned. Furthermore, we demonstrate that the experimental design of the breeder and the corresponding statistical assumptions should not be ignored in new AI methods, as large gains can be made from their integration. We expect closer collaboration between breeders and AI researchers to enable further gains, as both the experimental design and statistical models could be optimized for downstream predictive accuracy with neural networks.

\subsection{Comparison of methods}
MixINN, as a neural network-based approach, offers advantages over kernel-based methods through improved scalability and higher capacity to model nonlinear relationships. While kernel methods suffer from quadratic scaling in computational cost, neural networks continue improving with more data. However, we do note that G×EBLUP outperformed MixINN at identifying the $<5\%$ best genotypes per environment, i.e. G×EBLUP demonstrates better performance in recommending genotypes to specific farms, while MixINN demonstrates better overall performance at selection in breeding. To enhance the performance of MixINN at a farm-level, we suggest specialized recommendation losses that provide closer control over the top percentage to prioritize, such as the LambdaLoss framework~\cite{wang2018lambdaloss}.

The proposed method, MixINN, improves over other neural network based methods in ranking prediction by incorporating linear mixed models. This enables MixINN to separate genotype-specific, environment-specific and interaction effects from heterogeneous noise and generalize effectively to new genotypes and environments. 

A limitation of the MixINN two-step approach is that errors from the statistical model propagate to the neural network. Mitigating this issue would require end-to-end training that jointly models genetic- and spatiotemporal variance-covariance structures while learning feature mappings. We identify two promising approaches for this integration: deep neural networks with random effects~\cite{simchoni2023integrating} and Gaussian processes~\cite{rasmussen2003gaussian}. Integration of random effects into neural networks, as recently proposed by~\cite{simchoni2023integrating} enables a direct translation of the assumptions on VCOV structures to constraints on neural network optimization. Gaussian processes~\cite{rasmussen2003gaussian} capture complex VCOV structures through kernel functions and have previously been integrated with neural networks to model spatiotemporal correlation structures for yield forecasting~\cite{you2017deep}. 

\subsection{Reproducibility and facilitation of follow-up work}
In order to ensure reproducibility and facilitate follow-up work, we reorganized the used plant breeding dataset~\cite{lima2023genomes} into a reproducible benchmark by introducing domain-specific preprocessing, evaluation metrics and prediction scenarios. We extend previous results with implementations of established statistical methods and an additional neural network-based baseline. Upon publication, we will make the data, baseline models and modeling results for SINN and MixINN publicly available. We will share the source code for the processed benchmark dataset, including benchmark models, splits and metrics as supplementary material, described in Appendix D. Furthermore, we include a specification of the computing infrastructure and software in Appendix E.

\subsection{Limitations}
We recognize that the success of environment-specific selection is conditional on the accuracy of predicting future environmental conditions. However, we argue that seasonal weather prediction constitutes its own distinct application that is increasingly targeted by modern AI methods~\cite{nguyen2023climax}. This work demonstrates that environment-specific selection is challenging, even when the environmental conditions are known, and guides the development of models that address this challenge. We expect plant breeding to benefit from future work in seasonal forecasting, and encourage follow-up work integrating both challenges.

\subsection{Broader impact}
The demonstrated methodology improves crop yield under future climate conditions without relying on any crop- or region-specific assumptions. Predictive methods in plant breeding have historically found success across diverse crops, traits and regions~\cite{alemu2024genomic}. As such, improvements in this benchmark dataset and task can be expected to support improved prediction in breeding programs across crops and regions, as long as the breeding program is sufficiently large-scale. Additionally, selection is effective for any measurable and heritable crop trait, supporting impact beyond food security. Improved effectiveness of breeding programs can also reduce the environmental footprint of agriculture, through reduced need for pesticides~\cite{robinson1996return}, fertilizer~\cite{lammerts2017diverse}, and  water~\cite{condon2004breeding}.

\section{Conclusion}
This work demonstrates that combining statistical models for plant breeding with deep learning models for interaction prediction directly contributes to future food security, by improving ranking of genotypes and estimates of crop yields in new environmental conditions. We integrate modern methods for quantitative genetics with neural networks into a novel approach named MixINN, which integrates factor-analytic approximation of variance-covariance structures with a two-tower neural network model. MixINN is, to our knowledge, the first neural network-based approach to outperform domain-specific statistical methods in ranking accuracy of genotypes in new environments, surpassing state-of-the-art performance on open plant breeding data. Crucially, we show that these improvements in ranking lead to increased crop yields when MixINN is used to inform selection of genotypes. Our contribution enables further development of methods for predicting environment-specific genotype rankings, by making a new real-world challenge accessible to the AI community. Further progress on this task can directly contribute to future food security and sustainable agriculture under climate change.

\section*{Acknowledgements}
This work was partially supported by the Horizon Europe project PHENET - Tools and methods for extended plant PHENotyping and EnviroTyping services of European Research Infrastructures (Grant agreement ID 101094587).

\bibliographystyle{named}
\bibliography{ijcai26}

\section{Appendix}

\subsection*{A. Dataset filtering and imputation}
In this section, we describe our preprocessing steps for the Genomes to Fields 2022 Maize G×E Prediction Challenge \cite{lima2023genomes} dataset, filtering of samples and imputation of features. The dataset was divided into two sets: a training set (years 2015-2021) and a test set (year 2022).

\subsubsection{Filtering of samples} We filtered out incomplete samples from the training set. Samples with missing plot yields, genotypes with missing genetic markers and environments with missing weather data were removed. This left 123,517 samples for 4,417 genotypes and 212 environments in the training data.

\subsubsection{Filtering of features} For all feature types, we removed all features with more than 30\% missing values. We removed specific genetic markers with a minor allele frequency below 1\% or with more than 10\% missing values. The resulting markers were downsampled at random to a set of of 20,000 marker features. Daily weather features were aligned to 7 days before sowing until 133 days after sowing, giving 140 daily weather features.

\subsubsection{Imputation of features}
All imputation was conducted using the training dataset. We implemented modality-specific imputation methods as follows: genetic features were imputed using the mode at each position. Management features were also imputed with by mode. Soil features were imputed with the nearest neighbor in space across training environments. Environments with no weather features were excluded. Time-steps of weather features missing within a sample were imputed by linear imputation.

\subsection*{B. Hyperparameter tuning}
We conducted hyperparameter tuning for each neural network-based model on a single validation fold. This fold contained all environments of the year 2021, and a random 12.5\% of training genotypes. The specific label sets and hyperparameter ranges varied per model. With the exception of the models trained on $y^e_j$, MSE was used as evaluation criterion and one replicate was used per setting. All models trained on $y^e_j$ exhibited unstable performance during early experiments, due to the small number of validation samples (27 environments). As such, we average the results over 5 replicates, and use a modified evaluation criterion of $MSE-5r$ to ensure stable performance of selected models. An overview of hyperparameter ranges is given in Table~\ref{tab:appendix:hyper}. From the given ranges, we sampled random configurations to tune. No early stopping was used.

G2F-DNN~\cite{kick2023yield} was trained on crop yields, and only validation samples with both unseen genotypes and unseen environments were used to evaluate each model. The architectural hyperparameters were kept fixed, as the original work reported extensive tuning on a subset of the same dataset. We optimized learning rate and weight decay over similar ranges as the other models. SINN~\cite{potze2025} and MixINN (this work) contained component models that were tuned on label sets $y^g_i$, $y^e_j$ and $y^{ge}_{ij}$. Furthermore, we included a small set of architectural hyperparameters in hyperparameter tuning.

\begin{table*}[!t]
\centering

\begin{tabular}{@{}llC{0.8cm}C{1.3cm}C{4cm}C{2.5cm}C{2.5cm}@{}}
\toprule
\textbf{Model} & \textbf{Target} & \textbf{\#} & \textbf{Layers} & \textbf{Nodes per Layer}  & \textbf{Learning Rate} & \textbf{Weight Decay} \\
\midrule
G2F-DNN & $y_{ijk}$ & 25 &  & & 1e-4,  3e-4, 1e-3, 3e-3, 1e-2 & 1e-4,  3e-4, 1e-3, 3e-3, 1e-2  \\
\midrule
SINN ($f_g$) & $y^{g}_{i}$ & 125 & & 64, 96, 128, 192, 256 & 1e-4, 3e-4, 1e-3, 3e-3, 1e-2 & 1e-4,  3e-4, 1e-3, 3e-3, 1e-2 \\
MixINN ($f_g$) & $y^{g}_{i}$ & 125 & & 64, 96, 128, 192, 256 & 1e-4, 3e-4, 1e-3, 3e-3, 1e-2 & 1e-4,  3e-4, 1e-3, 3e-3, 1e-2 \\
\midrule
SINN ($f_e$) & $y^{e}_{j}$ & 200×5 & 3, 4, 5 & 8, 16, 32, 48, 64 & 1e-5,  3e-5, 1e-4, 3e-4, 1e-3 & 1e-5,  5e-5, 1e-4, 3e-4, 1e-3  \\
MixINN ($f_e$) & $y^{e}_{j}$ & 200×5 & 3, 4, 5 & 8, 16, 32, 48, 64 & 1e-5,  3e-5, 1e-4, 3e-4, 1e-3 & 1e-5,  5e-5, 1e-4, 3e-4, 1e-3  \\
\midrule
SINN ($f_{ge}$) & $y^{ge}_{ij}$ & 125 &  & 8, 16, 32, 64, 128 & 1e-4,  3e-4, 1e-3, 3e-3, 1e-2 & 1e-4,  3e-4, 1e-3, 3e-3, 1e-2 \\
MixINN ($f_{ge}$) & $y^{ge}_{ij}$ & 125 &  & 8, 16, 32, 64, 128 & 1e-4,  3e-4, 1e-3, 3e-3, 1e-2 & 1e-4,  3e-4, 1e-3, 3e-3, 1e-2 \\

\bottomrule
\end{tabular}\caption{Details on hyperparameter selection}\label{tab:appendix:hyper}
\end{table*}

\subsection*{C. Statistical baselines}
In this section, we describe our implementation of baseline models GBLUP~\cite{bernardo1994prediction,meuwissen2001prediction} and G×EBLUP~\cite{jarquin2014reaction}. We define GBLUP as following:
\begin{subequations}
\begin{align}
        y_{ijk} &= \mu + g_i + \epsilon_{ijk}, \\
        g_i &\sim N(0, \sigma^2_g\boldsymbol{\Sigma}_g),\\\boldsymbol{\Sigma}_g &= \frac{x^g {x^g}^\top}{tr(x^g {x^g}^\top)/n_g},
    \end{align}
\end{subequations}
\noindent
where $\Sigma_g$ is the genomic relationship matrix, estimated from genetic features $x^g$. Similarly, we implement G×EBLUP as follows:

\begin{subequations}
\begin{align}
    y_{ijk} &= \mu + g_i + e_j + ge_{ij} + \epsilon_{ijk},\\
        g_i &\sim N(0, \sigma^2_g\boldsymbol{\Sigma}_g),\\
        e_j &\sim N(0, \sigma^2_e\boldsymbol{\Sigma}_e),\\
        ge_{ij} &\sim N(0, \sigma^2_{ge}\boldsymbol{\Sigma}_g\otimes\boldsymbol{\Sigma}_e),\\
        \boldsymbol{\Sigma}_g &= \frac{x^g {x^g}^\top}{tr(x^g {x^g}^\top)/n_g}, \\
        \boldsymbol{\Sigma}_e &= \frac{x^e {x^e}\top}{tr(x^e {x^e}^\top)/n_e},
    \end{align}
\end{subequations}
\noindent

where \mbox{$\boldsymbol{\Sigma}_g \in \mathbb{R}^{n_g \times n_g}$} and \mbox{$\boldsymbol{\Sigma}_e \in \mathbb{R}^{n_e \times n_e}$} represent genomic and environmental relationship matrices, respectively. Environmental features $x^e$ for each environment were created by concatenating flattened daily weather features, soil features and management features into a single vector. Identical sets of features were used in GBLUP, G×EBLUP, G2F-DNN, SINN and MixINN. GBLUP and G×EBLUP were fitted using restricted maximum likelihood~\cite{patterson1971recovery} and implemented in R library \textit{qgg}~\cite{rohde2020qgg}.

\subsection*{D. Source code}
Upon publication, we will release the source code needed to reproduce the main experiments in the Code and Data Appendix. It includes a full preprocessing pipeline for the dataset~\cite{lima2023genomes}, baseline models (GBLUP, G×EBLUP, G2F-DNN), baseline results (SINN, MixINN, competition results), standardized splits and metrics, and experimentation interface for deep learning-based models and kernel-based models.

\subsection*{E. Hardware and software}
All experiments were conducted on a single node in a high performance compute (HPC) cluster, running 64-bit Ubuntu 24.04.2 with Linux version 5.15.0-88-generic. The node was equipped with 512~GB of RAM, 32 CPU cores and 4 NVIDIA A100 GPUs. Job scheduling was handled using SLURM version 24.11.0. For Python, R and CUDA versions and a detailed list of all software libraries, we refer to the \texttt{README.md} and \texttt{requirements.txt} files included in the future Code and Data Appendix. Finally, we list the software that was used during this work, but omitted from the Code and Data Appendix as follows:
\begin{itemize}
  \item \textbf{comet\_ml} (version 3.34.0) -- Proprietary Python package and platform, used for hyperparameter tuning and experiment tracking~\cite{comet_ml}.
  \item \textbf{asreml-R} (version 4.2) -- Proprietary R library, used for fitting linear mixed models with factor-analytic covariance structures~\cite{butler2017asreml}.
\end{itemize}

\end{document}